\documentclass[tablecaption=bottom,wcp]{jmlr} 
\usepackage{graphicx}

\jmlryear{2017}
\jmlrvolume{1}
\jmlrworkshop{ICML 2017 AutoML Workshop}

\title[Dynamic Input Structure and Network Assembly \titlebreak \hspace{0.01mm} for Few-Shot Learning]{Dynamic Input Structure and Network Assembly \titlebreak  for Few-Shot Learning}
 
\author{\Name{Nathan Hilliard} \Email{nathan.hilliard@pnnl.gov} \\
\addr Seattle, WA. 98109. \\
	\Name{Nathan O. Hodas} \Email{nathan.hodas@pnnl.gov} \\
    \Name{Courtney D. Corley} \Email{court@pnnl.gov} \\
    \addr Richland, WA. 99352. 
}

\begin{document}

\maketitle

\begin{abstract}
The ability to learn from a small number of examples has been a difficult problem in machine learning since its inception.
While methods have succeeded with large amounts of training data, research has been underway in how to accomplish similar performance with fewer examples, known as \textit{one-shot} or more generally \textit{few-shot} learning.
This technique has been shown to have promising performance, but in practice requires fixed-size inputs making it impractical for production systems where class sizes can vary.  This impedes  training and the final utility of few-shot learning systems. 
This paper describes an approach to constructing and training a network that can handle arbitrary example sizes dynamically as the system is used.
\end{abstract}


\section{Introduction}
\label{sec:intro}
Few-shot learning aims to alleviate the difficulty of learning a classifier from few examples -- or even a single example. 
Traditional classification  learns from hundreds or thousands of examples of per class. 
Instead of hoping to learn a classifier that can look at an input and classify it directly, a more robust technique is to provide the network with examples of each class and have it explicitly compare the input to each of the reference objects in each class. 
This typically takes the form of learning representations for both the reference examples you show the network as well as the query input that you are ultimately classifying.
Something of a similarity metric between the representations is then either learned as described by \cite{lens-fsl} or an out-of-the-box technique is used more explicitly, as in \cite{matchingnets}. 
The reference class with the highest similarity metric to the query image would be the label.

However, a practical concern is that the networks generally require all inputs to have the same number of examples per reference class.  This is largely unrealistic and unworkable for production systems.  Each reference class could have a different number of images (for example, a dynamic number of images for each family member in a facial recognition system.) 
Dynamic input sizing remains a challenging problem  for high performance techniques that utilize statically compiled graphs, such as Tensorflow and Caffe.
Clever workarounds have thusly been developed for different learning tasks, such as masking in sequence-based learning where sequences of varied length are padded to a fixed length.
The network learns to  monitor for the padding, and it can act accordingly, allowing for variable length sequences to utilize the same network inputs. However, it must explicitly learn to ignore the padding, placing an extra burden on the training process.

Our contribution in this paper is a novel technique for a system that leverages dynamic network assembly using shared weights to provide batch-wise size agnosticism in a static graph, meaning the example size changes from batch to batch. 
We additionally describe a training regimen that can be used to train the network to generalize and maintain similar performance across example sizes.
We demonstrate the architecture's effectiveness on a 1-way classification benchmark and compare against fixed-size networks.  We show that our contribution produces significantly higher performance on test tasks than a traditional static class-size approach.

\section{Related Work}
Since its inception, few-shot learning techniques have been implemented with a variety of architectures and components. 
For example, \cite{matchingnets} implemented one such network using memory augmentation with attention.
This builds on advances made by other few-shot systems such as the one developed by \cite{siamese} which used a siamese network with two convolutional tails achieving good performance on the omniglot dataset by \cite{omniglot}.
Other architectures such as pairwise networks demonstrated by \cite{pairwise-osl} have been used successfully as well.
Additionally, researchers have also looked at meta-learning regimes for training few-shot learning networks such as in \cite{meta-fsl}.
The goal being to train an LSTM that can provide updates to the network weights to a network during a few-shot training regimen.

Other tools for developing neural networks such as Chainer\footnote{https://github.com/chainer/chainer} and PyTorch\footnote{https://github.com/pytorch/pytorch}, which both use dynamic graph construction can also be used to address this problem.  However, these tools are  primarily for research and not production (although not impossible).
Our approach differs in that we define a straightforward and useful way to reuse the weights in statically compiled graphs with a more production-ready library such as TensorFlow (\cite{tensorflow}), giving us boosts in predictive accuracy as well as utility in production-grade applications.

\section{Methods}
\label{sec:method}
\subsection{Architecture}

\begin{figure}
	\centering
	\includegraphics[scale=0.65]{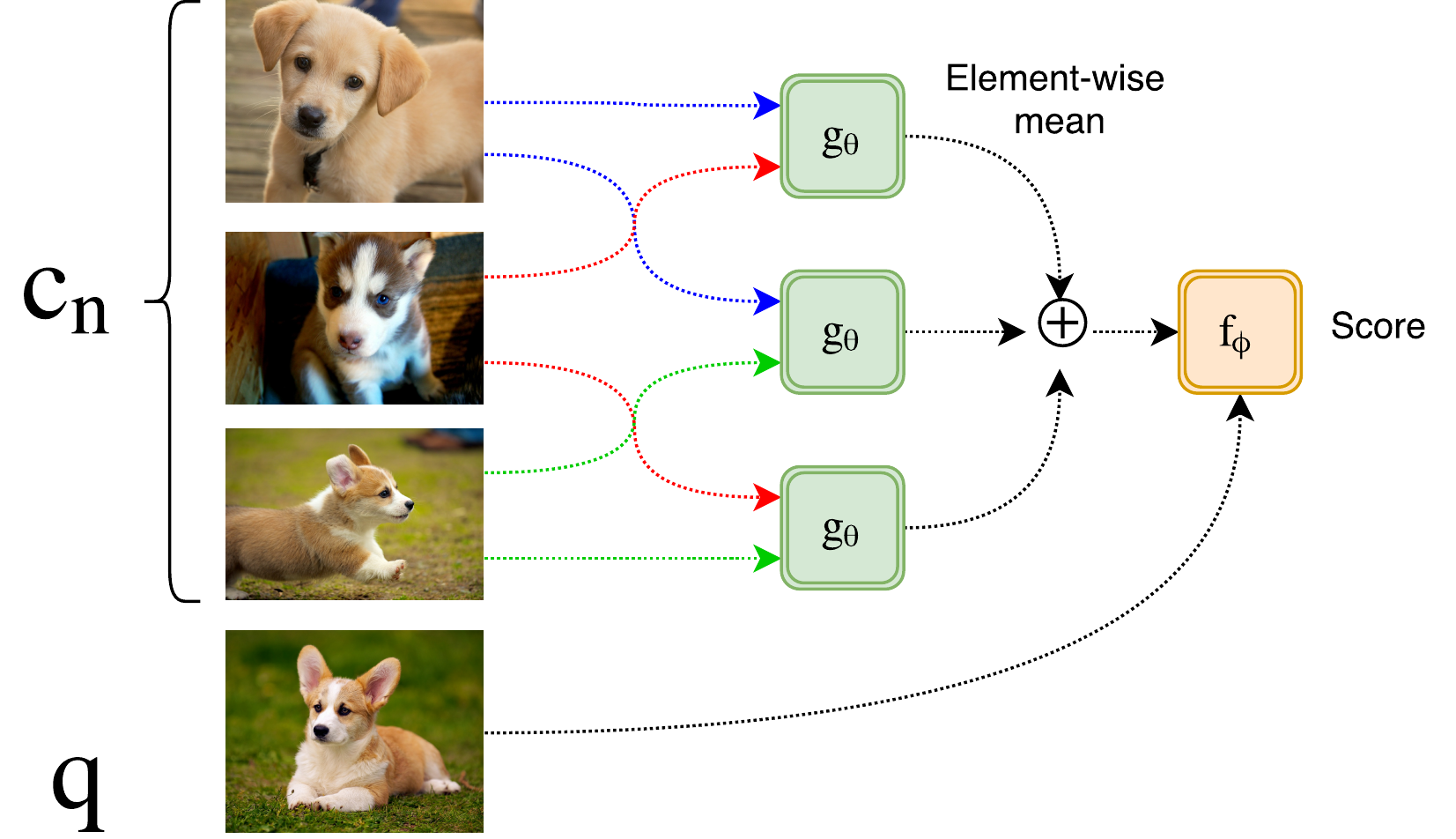}
    \vspace*{2mm}
    \caption{Our architecture, where $g_\theta$ and $f_\phi$ are both siamese networks. For simplicity, we represent $c_n$ and $q$ as images and omit the feature extraction step.}
    \label{fig:arch}
\end{figure}

We have a siamese two-stage network architecture where the first stage is based on work done by \cite{siamese} that leverages pairwise relational style networks most similar to work done by \cite{relationalnets} but also similar to \cite{pairwise-osl}.
The second stage of the network focuses on learning an internal distance metric, similar to metric learning as in \cite{metriclearning}.

To reduce dimensionality of our inputs and bootstrap our network, we use transfer learning described by \cite{transferlearning}.
Specifically, we extract features from the layer prior to the final classification layer from the residual network described in \cite{resnet}, this yields a vector of size 2048 in place of each image.
This also has several added benefits, namely we offload much of the feature learning to pre-trained networks.
Consequently, we then are able to instead focus each stage of our network on separate tasks.
Our architecture is similar to the work done by \cite{siamese} except our image features are extracted rather than learned.
A diagram visually describing our network can be seen in Figure~\ref{fig:arch}.

\subsubsection{Relational Stage}
First and foremost, we have a pairwise relational network $R$ that takes in a class $c$ with $n$ examples $c_n$. $R$ outputs a class embedding representing the set of examples $\{c_i\}_N$.

Ultimately, $R$ can be described as a function of the unique pairs (or \textit{combinations}) in $c_n$:

\begin{equation}
\label{eqn:R}
R(c_n) = \frac{1}{\binom{n}{2}} \sum\limits_{c_i, c_j}^{\binom{c_n}{2}} g_{\theta}(c_i, c_j)
\end{equation}

Where $g$ is a neural network parameterized by $\theta$, and $c_i, c_j$ are the embeddings of  $i$-th and $j$-th members of class $c$ where $i \neq j$. The $c$-embeddings could be provided by a pre-trained network or learned end-to-end.
It's important to note that while we use a similar pairwise comparison such as that used by \cite{relationalnets}, we crucially take the average vector from the resulting comparisons $g_\theta(c_i, c_j)$ instead of the sum of the comparisons. Averaging helps to enforce that the characteristic embedding of the class should not be explicitly related to the class size.
As such, we can use Equation~\ref{eqn:R} to learn an embedding describing that the class be explicitly considering the relationships between members of the class.

\subsubsection{Metric Stage}
The second stage network focuses on learning a distance metric between the query image $q$ and the given class  described by the output of $R(c_n)$.
Because the second stage and first stage are connected, we can learn better class embeddings via backpropogation.
In essence, this stage learns the probability that $q \in c_n$.

We describe the second stage network as $P$, defined as:
\begin{equation}
\label{eqn:P}
P(q, R(c_n)) = f_\phi(q, R(c_n)).
\end{equation}
Where $f$ is another neural network parameterized by $\phi$.

\subsection{Dynamic Assembly}
\begin{algorithm}
\floatconts
  {alg:assembly}%
  {\caption{Dynamic Assembly for Training \& Inferring on New Example Sizes}}%
{%
Given an input number of examples $n$, a feature vector size $s_v$
\begin{enumerate*}
  \item initialize a list $v$
  \item initialize an input tensor $input_c$ of size $[n, s_v]$
  \item initialize an input tensor $input_q$ of size $[s_v]$
  \item for each unique combination of indices $i, j \in \binom{n}{2}, i \neq j$:
  \begin{enumerate*}
    \item \label{step:create}Create a new instance of $g$ with shared weights $\theta$ i.e. $g_\theta(input_c[i], input_c[j])$, store the resulting output tensor in $v$.
  \end{enumerate*}
  \item store the concatenation of $v$ in an intermediary tensor $concat_v$.
  \item store the element-wise average of $concat_v$ in an intermediary tensor $avg_v$.
  \item connect $avg_v$ and $input_q$ to a new instance of $f$ reusing weights $\phi$ i.e. $f_\phi(input_q, avg_v)$
  \item \label{step:return} return $input_c, input_q,$ and  $f_\phi(input_q, avg_v)$
\end{enumerate*}
}%
\end{algorithm}

Because the reduction step in Equation~\ref{eqn:R} gives us a fixed sized vector, regardless of class size, we can use an arbitrary class size in the first stage.
We recreate $g_\theta$ for as many unique pairs that exist in $c$ dynamically, this is exemplified in Step~\ref{step:create} of Algorithm~\ref{alg:assembly}.
The result is that we ultimately only need to create intermediary operations between learned weights to accommodate new example sizes.
These operations along with an input for $q$ are then finally wired to $f_\phi$ in Equation~\ref{eqn:P} to complete the model creation.

Though we ultimately incur an overhead cost for creating these operations, it outweighs deployment cost of training separate networks for each example size and either storing them in memory simultaneously or swapping models every time class sizes change.
By storing the resulting input and output tensors from the creation, we can create an in-memory lookup table indexed by example size for each model based on the result of Step~\ref{step:return} in Algorithm~\ref{alg:assembly}.
This can then be easily incorporated in the batching step where the batcher is fed multiple classes each with varying example counts and returns that information in each batch so the appropriate inputs are used.

\section{Experiments}
\label{sec:exp}

\subsection{Experimental Design}
Our experiments were carried out by training the models end-to-end.
The baseline models are simply the same network trained on a fixed example size of training data whereas the dynamic input model was trained on varied example counts batch-to-batch.

For both cases, we trained via stochastic gradient descent with momentum with a learning rate of $\alpha = 0.001$ and momentum $\mu = 0.9$ with a batch size of $128$.
We used a final output layer with two units prior to a softmax activation function to determine whether $q \in c_n$ for a straightforward 1-way problem.

\subsection{Caltech-UCSD Birds}

\begin{table}[]
\centering
\begin{tabular}{llllll}
                        &                   &                   &                   &                   \\ \hline
\textbf{Training Class size:}     & \textbf{2}        & \textbf{3}        & \textbf{4}   & \textbf{5}   \\ \hline
\textbf{2-shot Network} & 62.3\%            & 62.5\%            & 62.7\%            & 62.7\%            \\
\textbf{3-shot Network} & 66.9\%            & 67.1\%            & 67.1\%            & 67.2\%            \\
\textbf{4-shot Network} & 71.9\%            & 72.2\%            & 72.3\%            & 72.3\%            \\
\textbf{5-shot Network} & 74.3\%            & 74.4\%            & 74.5\%            & 74.7\%            \\ \hline
\textbf{Dynamic Input}  & \textbf{89.7\%}   & \textbf{90\%}     & \textbf{90.1\%}   & \textbf{90.3\%}           
\end{tabular}
\caption{1-way Classification Results on the Caltech-UCSD Birds Dataset. Our experiments show fixed-size model performance on other example sizes in order to demonstrate how well they generalize. We create separate class sizes reusing the networks $g_\theta$ and $f_\phi$ that were trained on the fixed examples to evaluate the models on larger/smaller example sizes.}
\label{tbl:results}
\end{table}

Each model was trained on the same task using a dataset consisting of portions of the Caltech 256 and Visual Genome datasets developed by \cite{caltech256, visgenome}.
Similarly as mentioned earlier, for the feature extraction step for this experiment, we used the residual network built by \cite{resnet}.
We evaluate the models on a fine-grained unrelated classification task, the Caltech-UCSD Birds dataset described by \cite{birds}.
In order to train the dynamic model to better generalize across class sizes, we feed it random example sizes each batch.
More concisely, the first batch could contain 2-shot data with the next batch containing 5-shot data, and so on.
An indirect, yet important, contribution of this work is the observation that randomly changing class sizes significantly reduces overfitting.

In our experiments we consider baselines using the same architecture which are trained on a fixed example size.
Each model in Table~\ref{tbl:results} was trained with the same number of training steps.
As is shown by the table, our technique far out-performs networks trained on fixed example sizes.
The results show that even when trained on small example counts, when evaluating on higher example counts performance improves.
This is largely unsurprising, the more examples you show the network the better it should perform in general.
Networks trained on higher example counts outperforming networks trained on lower example counts is also an unsurprising result.
This is likely due in large part to the element-wise mean operation we use. 
As example counts get higher the resulting class vector should get sharper, making it easier for the network to make a distinction between it and the query image.

\subsection{Omniglot}

For the Omniglot dataset (\cite{omniglot}), we use a very simple network trained on MNIST digits as our feature extractor as opposed to ResNet.
All other experimental parameters are kept the same as that of the Caltech-Birds experiment except that we use Nesterov momentum instead of classic momentum as described by \cite{nesterov}.
Our model performs better than the baselines on this task by a narrower margin as can be seen in Table~\ref{tbl:omniglot}.

\begin{table}[]
\centering
\begin{tabular}{llllll}
                        &                   &                   &                   &                   \\ \hline
\textbf{Training Class size:}     & \textbf{2}        & \textbf{3}        & \textbf{4}   & \textbf{5}   \\ \hline
\textbf{2-shot Network} & 52.2\%            & 52.4\%            & 52.5\%            & 52.5\%            \\
\textbf{3-shot Network} & 80.5\%            & 81.4\%            & 81.6\%            & 81.8\%            \\
\textbf{4-shot Network} & 82.3\%            & 84\%              & 84.5\%            & 84.8\%            \\
\textbf{5-shot Network} & 82.8\%            & 84.4\%            & 85.2\%            & 85.7\%            \\ \hline
\textbf{Dynamic Input}  & \textbf{83.1\%}   & \textbf{84.9\%}     & \textbf{85.8\%}   & \textbf{86.2\%}           
\end{tabular}
\caption{1-way Classification Results on the Omniglot dataset.}
\label{tbl:omniglot}
\end{table}

\section{Conclusion}
\label{sec:end}
In this paper we presented a technique for bringing few-shot learning into the dynamic setting necessary for production applications.
By generalizing a network to multiple example sizes, a single network can perform few-shot classification on varied example sizes at runtime with a comparatively minimal overhead incurred.
We demonstrate that our dynamic model can perform much better than its fixed-size counterparts on a fine-grained task unseen during training time.

\newpage

\bibliography{automl17}

\begin{thebibliography}{15}
\providecommand{\natexlab}[1]{#1}
\providecommand{\url}[1]{\texttt{#1}}
\expandafter\ifx\csname urlstyle\endcsname\relax
  \providecommand{\doi}[1]{doi: #1}\else
  \providecommand{\doi}{doi: \begingroup \urlstyle{rm}\Url}\fi

\bibitem[Abadi et~al.(2015)Abadi, Agarwal, Barham, Brevdo, Chen, Citro,
  Corrado, Davis, Dean, Devin, Ghemawat, Goodfellow, Harp, Irving, Isard, Jia,
  Jozefowicz, Kaiser, Kudlur, Levenberg, Man\'{e}, Monga, Moore, Murray, Olah,
  Schuster, Shlens, Steiner, Sutskever, Talwar, Tucker, Vanhoucke, Vasudevan,
  Vi\'{e}gas, Vinyals, Warden, Wattenberg, Wicke, Yu, and Zheng]{tensorflow}
Mart\'{\i}n Abadi, Ashish Agarwal, Paul Barham, Eugene Brevdo, Zhifeng Chen,
  Craig Citro, Greg~S. Corrado, Andy Davis, Jeffrey Dean, Matthieu Devin,
  Sanjay Ghemawat, Ian Goodfellow, Andrew Harp, Geoffrey Irving, Michael Isard,
  Yangqing Jia, Rafal Jozefowicz, Lukasz Kaiser, Manjunath Kudlur, Josh
  Levenberg, Dan Man\'{e}, Rajat Monga, Sherry Moore, Derek Murray, Chris Olah,
  Mike Schuster, Jonathon Shlens, Benoit Steiner, Ilya Sutskever, Kunal Talwar,
  Paul Tucker, Vincent Vanhoucke, Vijay Vasudevan, Fernanda Vi\'{e}gas, Oriol
  Vinyals, Pete Warden, Martin Wattenberg, Martin Wicke, Yuan Yu, and Xiaoqiang
  Zheng.
\newblock {TensorFlow}: Large-scale machine learning on heterogeneous systems,
  2015.
\newblock URL \url{http://tensorflow.org/}.
\newblock Software available from tensorflow.org.

\bibitem[Bellet et~al.(2013)Bellet, Habrard, and Sebban]{metriclearning}
Aur{\'{e}}lien Bellet, Amaury Habrard, and Marc Sebban.
\newblock A survey on metric learning for feature vectors and structured data.
\newblock \emph{CoRR}, abs/1306.6709, 2013.
\newblock URL \url{http://arxiv.org/abs/1306.6709}.

\bibitem[Bengio et~al.(2011)]{transferlearning}
Yoshua Bengio et~al.
\newblock Deep learning of representations for unsupervised and transfer
  learning.
\newblock \emph{JMLR W\&CP: Proc. Unsupervised and Transfer Learning}, 2011.

\bibitem[Griffin et~al.(2007)Griffin, Holub, and Perona]{caltech256}
G.~Griffin, A.~Holub, and P.~Perona.
\newblock Caltech-256 object category dataset.
\newblock Technical Report 7694, California Institute of Technology, 2007.
\newblock URL \url{http://authors.library.caltech.edu/7694}.

\bibitem[He et~al.(2015)He, Zhang, Ren, and Sun]{resnet}
Kaiming He, Xiangyu Zhang, Shaoqing Ren, and Jian Sun.
\newblock Deep residual learning for image recognition.
\newblock \emph{CoRR}, abs/1512.03385, 2015.
\newblock URL \url{http://arxiv.org/abs/1512.03385}.

\bibitem[Koch(2015)]{siamese}
Gregory Koch.
\newblock \emph{Siamese neural networks for one-shot image recognition}.
\newblock PhD thesis, University of Toronto, 2015.

\bibitem[Krishna et~al.(2016)Krishna, Zhu, Groth, Johnson, Hata, Kravitz, Chen,
  Kalanditis, Li, Shamma, Bernstein, and Fei-Fei]{visgenome}
Ranjay Krishna, Yuke Zhu, Oliver Groth, Justin Johnson, Kenji Hata, Joshua
  Kravitz, Stephanie Chen, Yannis Kalanditis, Li-Jia Li, David~A Shamma,
  Michael Bernstein, and Li~Fei-Fei.
\newblock Visual genome: Connecting language and vision using crowdsourced
  dense image annotations.
\newblock 2016.

\bibitem[Lake et~al.(2015)Lake, Salakhutdinov, and Tenenbaum]{omniglot}
Brenden~M. Lake, Ruslan Salakhutdinov, and Joshua~B. Tenenbaum.
\newblock Human-level concept learning through probabilistic program induction.
\newblock \emph{Science}, 350\penalty0 (6266):\penalty0 1332--1338, 2015.
\newblock ISSN 0036-8075.
\newblock \doi{10.1126/science.aab3050}.
\newblock URL \url{http://science.sciencemag.org/content/350/6266/1332}.

\bibitem[Mehrotra and Dukkipati(2017)]{pairwise-osl}
Akshay Mehrotra and Ambedkar Dukkipati.
\newblock Generative adversarial residual pairwise networks for one shot
  learning.
\newblock \emph{CoRR}, abs/1703.08033, 2017.
\newblock URL \url{http://arxiv.org/abs/1703.08033}.

\bibitem[Ravi and Larochelle(2016)]{meta-fsl}
Sachin Ravi and Hugo Larochelle.
\newblock Optimization as a model for few-shot learning.
\newblock 2016.

\bibitem[{Santoro} et~al.(2017){Santoro}, {Raposo}, {Barrett}, {Malinowski},
  {Pascanu}, {Battaglia}, and {Lillicrap}]{relationalnets}
A.~{Santoro}, D.~{Raposo}, D.~G.~T. {Barrett}, M.~{Malinowski}, R.~{Pascanu},
  P.~{Battaglia}, and T.~{Lillicrap}.
\newblock {A simple neural network module for relational reasoning}.
\newblock \emph{ArXiv pre-prints arXiv:1706.01427}, June 2017.

\bibitem[Sutskever et~al.(2013)Sutskever, Martens, Dahl, and Hinton]{nesterov}
Ilya Sutskever, James Martens, George Dahl, and Geoffrey Hinton.
\newblock On the importance of initialization and momentum in deep learning.
\newblock In \emph{International conference on machine learning}, pages
  1139--1147, 2013.

\bibitem[{Triantafillou} et~al.(2017){Triantafillou}, {Zemel}, and
  {Urtasun}]{lens-fsl}
E.~{Triantafillou}, R.~{Zemel}, and R.~{Urtasun}.
\newblock {Few-Shot Learning Through an Information Retrieval Lens}.
\newblock \emph{ArXiv e-prints}, July 2017.

\bibitem[Vinyals et~al.(2016)Vinyals, Blundell, Lillicrap, Kavukcuoglu, and
  Wierstra]{matchingnets}
Oriol Vinyals, Charles Blundell, Timothy~P. Lillicrap, Koray Kavukcuoglu, and
  Daan Wierstra.
\newblock Matching networks for one shot learning.
\newblock \emph{CoRR}, abs/1606.04080, 2016.
\newblock URL \url{http://arxiv.org/abs/1606.04080}.

\bibitem[Welinder et~al.(2010)Welinder, Branson, Mita, Wah, Schroff, Belongie,
  and Perona]{birds}
P.~Welinder, S.~Branson, T.~Mita, C.~Wah, F.~Schroff, S.~Belongie, and
  P.~Perona.
\newblock {Caltech-UCSD Birds 200}.
\newblock Technical Report CNS-TR-2010-001, California Institute of Technology,
  2010.

\end{thebibliography}

\end{document}